\pdfoutput=1

\documentclass[11pt]{article}

\usepackage{ACL2023}

\usepackage{times}
\usepackage{latexsym}
\usepackage{hyperref}
\usepackage[T1]{fontenc}

\usepackage[utf8]{inputenc}

\usepackage{microtype}
\usepackage{tcolorbox}
\usepackage{xcolor}
\usepackage{enumitem}
\tcbuselibrary{skins}
\usepackage{makecell}
\usepackage{tabularx}
\usepackage{multirow}
\usepackage{booktabs}
\usepackage{fancyvrb}
\usepackage{siunitx}
\usepackage{longtable}
\usepackage[inkscapeformat=png]{svg}
\usepackage{svg}

\usepackage{inconsolata}

\definecolor{mygreen}{RGB}{100, 128, 100} 
\definecolor{fontcolor}{RGB}{255, 255, 255} 
\title{Generating Harder Cross-document Event Coreference Resolution Datasets using Metaphoric Paraphrasing}

\author{Shafiuddin Rehan Ahmed$^{1}$ \hspace{3mm} Zhiyong Eric Wang$^{2}$ \hspace{3mm} George Arthur Baker$^{1}$
\\[1.5mm]
\textbf{Kevin Stowe$^{3}$}  \hspace{3mm} \textbf{James H. Martin$^{1}$} 
\\[1.mm]Departments of $^{1}$Computer Science \& $^{2}$CLASIC, University of Colorado, Boulder, USA \\ \tt{\{shah7567, zhwa3087\}@colorado.edu}
\\[1.mm] $^{3}$Education Testing Service (ETS)}

\newtcolorbox{mybox}[1][]{enhanced,colback=green!5!white,
colbacktitle=green!40!black!70!white,
colframe=green!35!black,fonttitle=\small\bfseries,
underlay={\begin{tcbclipinterior}
\end{tcbclipinterior}},
attach boxed title to top center={yshift=-2mm,xshift=-10mm},#1}

\newcommand{\LH}{\ensuremath{\texttt{LH}}}
\newcommand{\devSmall}{\ensuremath{\tt{Dev}_{small}}}
\newcommand{\CE}{\ensuremath{\texttt{CE}}}
\newcommand{\CELH}{\ensuremath{\texttt{CE}_\LH}}
\newcommand{\CEKnn}{\ensuremath{\texttt{CE}_{\tt KNN}}}
\newcommand{\LLMLH}{\ensuremath{\texttt{GPT}_\LH}}
\newcommand{\LLMKnn}{\ensuremath{\texttt{GPT}_{\tt KNN}}}
\newcommand{\ecb}{{\tt ECB+}}
\newcommand{\ecbM}{{\tt ECB+META}}
\newcommand{\ecbMone}{{\tt ECB+META$_1$}}
\newcommand{\ecbMm}{{\tt ECB+META$_m$}}

\begin{document}
\maketitle
\begin{abstract}
%
%
The most widely used Cross-Document Event Coreference Resolution (CDEC) datasets fail to convey the true difficulty of the task, due to the lack of lexical diversity between coreferring event triggers (words or phrases that refer to an event). Furthermore, there is a dearth of event datasets for figurative language, limiting a crucial avenue of research in event comprehension. We address these two issues by introducing \ecbM, a lexically rich variant of Event Coref Bank Plus (\ecb) for CDEC on figurative and metaphoric language. We use GPT-4 as a tool for the metaphoric transformation of sentences in the documents of \ecb, then tag the original event triggers in the transformed sentences in a semi-automated manner. In this way, we avoid the re-annotation of expensive coreference links. We present results that show existing methods that work well on \ecb~struggle with \ecbM, thereby paving the way for CDEC research on a much more challenging dataset.\footnote{ Code/data:
\href{https://github.com/ahmeshaf/llms_coref}{github.com/ahmeshaf/llms\_coref}}
\end{abstract}

\section{Introduction}
\begin{figure*}[t]
    \centering
    \includegraphics[scale=0.73]{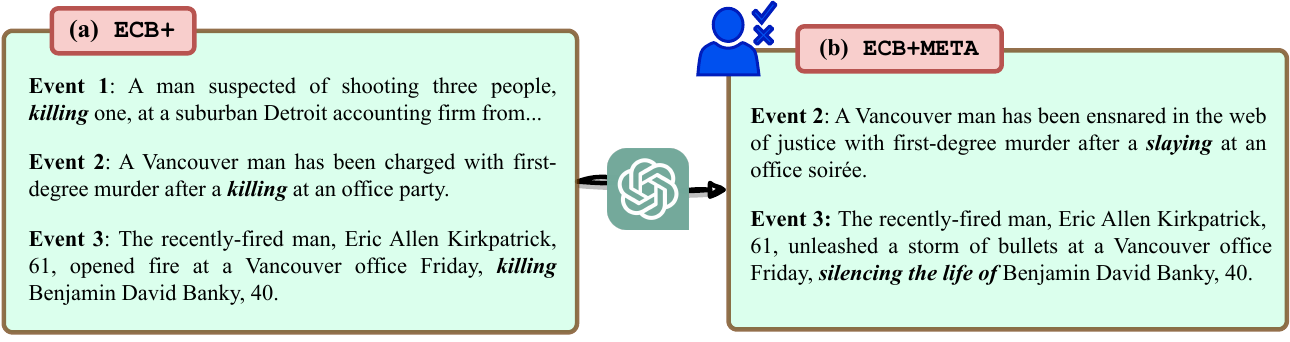}
    \vspace{-5.5mm}
    \caption{Using GPT-4 to Generate \ecbM~from \ecb Corpus. \textbf{Event 2} \& \textbf{Event 3} are coreferent, while \textbf{Event 1} is not. \ecbM~has metaphorically transformed triggers, e.g., \textit{killing} -> \textit{silencing the life}. The triggers are hand-corrected by an annotator. \ecbM~challenges previous work---\citet{held-etal-2021-focus} \& \citet{ahmed-etal-2023-2}. 
    \vspace*{-5.5mm}}
    \label{fig:pipeline}
\end{figure*}

Cross-Document Event Coreference Resolution (CDEC) involves identifying mentions of the same event within and across documents. 
An issue with CDEC is that the widely used dataset, Event Coref Bank plus (\ecb; \citet{cybulska-vossen-2014-using}), is  biased towards lexical similarities, both for triggers and associated event arguments, and therefore has a very strong  baseline \cite{cybulska-vossen-2015-translating, kenyon-dean-etal-2018-resolving,ahmed-etal-2023-2}. 
To see this, consider the excerpts from \ecb~shown in Figure \ref{fig:pipeline}(a). This consists of three \textit{killing} events selected from separate articles sharing a common trigger.  An algorithm capable of matching the triggers and tokens within the sentences, such as "\textit{Vancouver}" and "\textit{office}," can readily discern that Event 2 is coreferent with Event 3, and not Event 1. This leads  to the question of whether the state-of-the-art methods using this corpus \cite{held-etal-2021-focus} learn the semantics of event coreference, or are merely exploiting  surface triggers.

Figurative language, encompassing metaphors, similes, idioms, and other non-literal expressions, is an effective tool for assessing comprehension across cognitive, linguistic, and social dimensions \cite{lakoff1980metaphors, 
winner1988point, gibbs1994poetics, palmer2004reading, palmer2006bridging}. Figurative language, by its nature, draws on a wide array of cultural, contextual, and imaginative resources to convey meanings in nuanced and often novel ways. Consequently, it employs a broader vocabulary and more unique word combinations than literal language \cite{stefanowitsch2006words}.
Most recent work on metaphors has been focused on generation \cite{stowe2020metaphoric, stowe-etal-2021-metaphor, chakrabarty2021mermaid}, interpretation \cite{chakrabarty2022flute, chakrabarty-etal-2023-spy}, and detection \cite{li-etal-2023-framebert,joseph-etal-2023-newsmet, wachowiak2023does}. 
Yet, there is a dearth of event datasets for figurative language which limits an important research direction of event comprehension.

In this paper, we address these two challenges by leveraging GPT-4 in \textit{constrained metaphoric paraphrasing} of \ecb documents. We introduce a novel dataset named \ecbM~, which we generate using a semi-automatic approach. This involves applying metaphoric transformations to the event triggers within \ecb~and then hand-correcting the tagged triggers in the new corpus. As depicted in Figure \ref{fig:pipeline}(b), the trigger word \textit{killing} in Events 2 and 3 of \ecb~become \textit{slaying} and \textit{snuffing out the flame of life of} in \ecbM, respectively.

This approach preserves the coreference annotations from \ecb, thereby avoiding an expensive coreference re-annotation task. Thus, we create several versions of ``tougher'' CDEC benchmark datasets with enhanced lexical diversity with varying levels of metaphoricity. We present baseline results using previous methods---\citet{held-etal-2021-focus} and \citet{ahmed-etal-2023-2} (described in \S\ref{sec:cdec}), and show the limitation of these approaches on this dataset. Finally, we correlate lexical diversity and text complexity with CDEC and test the hypothesis that CDEC gets more difficult as the lexical diversity/complexity of the corpus increases.

\section{Related Work}
\vspace{-0.5mm}
\subsection{CDEC Datasets}
\vspace{-0.4mm}
\ecb\footnote{Corpus detailed in \S\ref{sec:app-ecb}} is the most widely used dataset for CDEC, yet it has limited utility in realistic applications because of how simple the dataset is. The Gun Violence Corpus (GVC; \citet{vossen-etal-2018-dont}), for instance, was introduced as a way of adding ambiguity to the task. Yet, both these datasets lack lexical diversity in terms of coreferent event triggers. \citet{ravenscroft-etal-2021-cd} is one such work that addresses the diversity question through cross-domain coreference, however, a dataset focusing CDEC on figurative language does not exist to our best knowledge.

Even with the use of modern annotation tools \cite{klie-etal-2018-inception, ahmed-etal-2023-good}, annotating CDEC datasets is expensive.
Works such as \citet{bugert-gurevych-2021-event, eirew-etal-2021-wec} use Wikipedia as a way of bootstrapping ECR annotations automatically. In a similar vein, we bootstrap CDEC annotations for figurative language in a synthetic way using GPT-4.

\newcommand{\BEKnn}{\ensuremath{\mathtt{KNN}}}
\vspace{-0.3mm}
\subsection{Metaphoric Paraphrasing}
\vspace{-0.3mm}
The task of metaphoric paraphrasing has been explored through a variety of methods. A primary theme is sentential paraphrasing by replacing literal words with metaphors \cite{stowe-etal-2021-exploring,stowe-etal-2021-metaphor,chakrabarty-etal-2021-mermaid}. These approaches fine tune language models with control codes to indicate metaphors, exploiting available metaphoric data to facilitate transformations from literal language to metaphoric. However, they rely on extensive data, and there is evidence that modern large language models excel at metaphor generation \cite{chakrabarty-etal-2023-spy} and paraphrasing \cite{kojima2023large, openai2023gpt4}. For this reason, we leverage GPT-4 via ChatGPT functionality for our experiments.

\vspace{-0.1mm}
\subsection{CDEC Methods}

\noindent \textbf{Non-filtering Methods:}
Previous works \cite{meged-etal-2020-paraphrasing,zeng-etal-2020-event,
cattan-etal-2021-cross-document, allaway-etal-2021-sequential, caciularu-etal-2021-cdlm-cross,yu-etal-2022-pairwise} in CDEC   have been successful using pairwise mention representation learning models, a method popularly known as cross-encoding. These methods use distributed and contextually-enriched ``non-static'' vector representations of mentions from Transformer-based \cite{NIPS2017_3f5ee243} language models like various BERT-variants \cite{devlin-etal-2019-bert, Beltagy2020Longformer} to calculate supervised pairwise scores for those event mentions. While these methods demonstrate SoTA performance, their applicability is hindered by their quadratic complexity at inference.

\vspace{1mm}
\noindent \textbf{Filtering Methods:}
Keeping usability and tractability in mind, we experiment only with the recent work that adds a low-compute mention pair filtering step before crossencoding. These approaches aid in the removal of numerous irrelevant mention pairs, thereby directing focus toward the most pertinent pairs with resource-intensive models. For instance, in their work, \citet{held-etal-2021-focus} propose a retrieval, vector-based K-nearest neighbor method, that helps find and focus only on the hard negatives in the corpus. In contrast, \citet{ahmed-etal-2023-2} employ simplified lexical similarity metrics to filter out a substantial number of truly non-coreferent pairs in the corpus.




\section{Methodology}
\vspace{-1mm}
We first synthetically create \ecbM~ by employing metaphoric paraphrasing of the original corpus. Then we tag the event triggers of the original corpus in \ecbM~ in a semi-automated manner. Finally, we adopt two existing CDEC methods to test this new dataset. We describe each of these steps:
\vspace{-0.5mm}
\subsection{Metaphoric Paraphrasing using GPT-4}
 We paraphrase \ecb's sentences in a constrained manner in which we convert only the event triggers in a sentence into metaphors. We first extract the event mentions from each sentence of the documents in the corpus, then prompt GPT-4 to convert only the trigger words in the sentence to metaphors. We adopt a chain of thought prompting approach \cite{kojima2022large}, where we provide the steps that need to be followed in the conversion (see \S\ref{sec:app:metaprompt}).

To enhance diversity and sample appropriate metaphors, we generate five metaphors for every trigger word in the sentence and then task GPT-4 to select the most coherent one from the list. We diversify metaphoricity levels by using both single-word and multi-word metaphors. As illustrated in Figure \ref{fig:paraphrase}, the conversion of "\textit{killing}" into a single-word metaphor is "\textit{slaying}," while its transformation into a multi-word phrasal metaphor is "\textit{extinguishing the candle of life}." We develop two versions of \ecbM, designated as \ecbMone~for single-word transformations and \ecbMm~for multi-word transformations, respectively.

Using the generated conversions, we first automatically tag the original events in the transformed sentences. Then, we hand-correct cases where the conversion is ambiguous. In the end, we are left with two versions of the validation and the test sets of \ecbM~preserving the original coreference annotations of \ecb.


\vspace{-0.5mm}
\subsection{CDEC Methods}
\label{sec:cdec}
\paragraph{Filtering Step for CDEC:}
The BiEncoder K-NN (\BEKnn) approach, introduced by \citet{held-etal-2021-focus} involves a novel approach to mention pair retrieval before doing CDEC. This method focuses on selecting mentions that are most similar to a given target mention using their static vector representations and a Vector Store (like FAISS \citet{johnson2019billion}). To achieve this, they fine-tune the RoBERTa-Large \cite{liu2019roberta} pre-trained model using a contrastive Categorical Loss function, with categories corresponding to event clusters within the corpus. This fine-tuning process utilizes token embeddings generated by the language model and trains on the centroid representations of gold standard event clusters. Due to computation constraints, we use RoBERTa-Base instead of RoBERTa-Large in this work. For the same reason, we use triplet-loss with mention pairs instead of the centroid of clusters.

The Lemma Heuristic (\LH; \citet{ahmed-etal-2023-2})  leverages lexical features to pre-filter non-coreferent pairs before CDEC. This way, they eliminate the need for an additional fine-tuning step as required in the \BEKnn~approach. \LH~focuses on creating a balanced set of coreferent and non-coreferent pairs while minimizing the inadvertent exclusion of coreferent pairs (false negatives) by the heuristic. It accomplishes this by first generating a set of synonymous lemma pairs from the training corpus and then applying a sentence-level word overlap ratio to prune pairs that don't meet the threshold or lack synonymy. In this work, we use the \LH~method for filtering and also as a baseline lexical method following \citet{ahmed-etal-2023-2}.

\paragraph{Cross-encoder\footnote{Described in more detail in \S\ref{sec:app-cross}}:} The Cross-Encoder (CE) functions within CDEC as a pairwise classifier, leveraging joint representations of a mention pair \((e_i, e_j)\). First, it combines the two event mentions with their respective contexts into a single unified string to facilitate cross-attention. Next, it derives the token-level representations of each mention after encoding this unified string. Finally, the joint representation is the concatenation of the context-enhanced token representations \((v_{e_i}, v_{e_j})\) along with their element-wise product, as illustrated below:

\vspace{-2mm}
\begin{equation}
    v_{(e_i, e_j)} = [v_{e_i} , v_{e_j} , v_{e_i} \odot v_{e_j} ]
\end{equation}
The resulting vector \(v_{(e_i, e_j)}\) is then refined through a binary cross-entropy loss function using logistic regression that learns coreference. 
In our work, we use the learned weights of the \CELH\footnote{Provided by the authors}. For the \BEKnn~cross-encoder (\CEKnn), we trained the weights of RoBERTA-Base using the \BEKnn~to generate focused mention pairs. We carry out our experiments in a transfer learning format where we train the crossencoders only on the training set of \ecb~and use the test sets of \ecbM. This is motivated by the work of \citet{ORTONY1978465}, which argues the human processes required for comprehension of figurative and literal uses of language are essentially similar.
\paragraph{GPT-4 as Pairwise Classifier:} 
\citet{yang2022gpt} demonstrated the viability of a prompt-based binary coreference classifier using GPT-2, though the results were sub-par. Building on their work, we employ a similar prompting technique with GPT-4 to develop an enhanced classifier. This classifier determines whether a pair of events, identified by marked triggers in sentences, are coreferent by responding with ``Yes'' or ``No''. Similar to \CE, we vary this method by incorporating the two fitering techniques (\LLMLH, \LLMKnn)
\newcommand{\conll}{{\small CoNLL}}

\newcommand{\bertscore}{\ensuremath{\mathtt{BS}}}
\newcommand{\BSKnn}{\ensuremath{\bertscore}}
\label{sec:heuristics}

\vspace{1.5mm}
\newcommand{\BcuR}{\normalsize \ensuremath{\textsc{B}^{3}_{\textsc{R}}}}
\newcommand{\BcuP}{\normalsize \ensuremath{\textsc{B}^{3}_{\textsc{P}}}}
\newcommand{\BcuF}{\normalsize \ensuremath{\textsc{B}^{3}_{\textsc{F1}}}}
\newcommand{\Bcu}{\normalsize \ensuremath{\textsc{B}^{3}_{}}}

\vspace*{-1.5mm}
\section{Results}
\vspace*{-0.3mm}
\subsection{Metaphor Quality Control}
To assess the quality of the generated metaphors, an annotator familiar with the events in the \ecb~dataset manually examines the \devSmall~sets. We chose a familiarized annotator because metaphors often abstract away many of the details that make coreference obvious, and we are interested in whether or not the generated paraphrases would (by any stretch of the imagination) reasonably be interpreted as referring to the original event.

The annotator examines each of the original event mentions alongside their paraphrased versions and makes a binary judgment as to whether the two can be reasonably interpreted as referring to the same event. We estimate based on the results that approximately 99\% of \ecbMone~and 95\% of \ecbMm~could be reasonably interpreted by a human as being coreferent to the original event mentions from which they are derived.

\subsection{Coreference \& Lexical Diversity}
We use \Bcu~\cite{bagga1998algorithms} and {\conll} \cite{denis2009global, pradhan-etal-2012-conll} clustering metrics, in which we use the \BcuR~for estimating recall, {\conll} as the overall metric (evaluated using CoVal \cite{moosavi-etal-2019-using}). For the methods that use \LH~as the filtering step, we follow \citet{ahmed-etal-2023-2}'s clustering with connected components. For \BEKnn~as the filtering step,  we use \citet{held-etal-2021-focus}'s greedy agglomeration.



\paragraph{Filtering Scores:} Following previous work, we first assess the \BcuR~score on oracle results. This tests how well the filtering methods perform in minimizing false negatives (coreferent pairs that are eliminated inadvertently). From Table \ref{tab:ecb-Meta-heu-recall} we observe a substantial difference in the recall measures of \ecb~and \ecbM~versions. The \LH~approach particularly takes a toll because it relies on synonymous lemma pairs from the train set. Interestingly, \BEKnn~does well on the \ecbM~versions, with only a minor drop in recall for \ecbMone~and about 10\% drop for \ecbMm. Between \ecbMone~and \ecbMm, as expected, the recall drops more in \ecbMm~as more complex metaphors are used here.

\renewcommand{\arraystretch}{1.3}
\begin{table}[t!]
\small
\centering
    \begin{tabular}{cc@{}|rcccc|c@{}}
        \toprule
       \multicolumn{2}{c}{~}& \textbf{Method} && \textbf{Dev}  & \textbf{\devSmall} & \multicolumn{1}{c}{\textbf{Test}} & \\ 
 \hline
 \multirow{2}{*}{\makecell{\ecb}} &&
 \LH &&76.3 & 87.9 & 81.5 &\\
 && \BEKnn &&95.7 & 95.3 & 94.9 &\\
\cline{1-8}
\multirow{2}{*}{\makecell{ \tt ECB+\\ \tt META$_1$  }} &  & \LH &&45.8 & 64.6 & 58.2  &\\
&& \BEKnn &&91.8& 93.7 & 91.4  &\\
\cline{1-8}
\multirow{2}{*}{\makecell{\tt ECB+\\ \tt META$_m$}} &  & \LH &&38.4 & 59.4 & 51.3 &\\
&& \BEKnn &&84.4& 86.5 & 85.6  &\\
\cline{1-8}
\bottomrule
    \end{tabular}
  \caption[ECB+ Results]{\BcuR~Oracle Results on Dev, \devSmall~and Test sets of \ecb, \ecbMone, and \ecbMm.  
	}
\label{tab:ecb-Meta-heu-recall}
\vspace*{-3mm}
\end{table}

 \renewcommand{\arraystretch}{1.3}
\begin{table}[h!]
\centering
\small
    \begin{tabular}{c@{}|rccc|c@{}}
        \toprule
        \multicolumn{1}{c}{~}&\textbf{Method} & ECB+ & \makecell{ECB+\\META$_1$}&  \multicolumn{1}{c}{\makecell{ECB+\\META$_m$}}& \\
 \hline
& \LH& 74.1 &  49.8 & 54.0 &\\
&\CELH & 78.1 & 60.9 & 50.6 & \\
 &\CEKnn &  78 & \textbf{71.4}&  54.8&\\
 &\LLMLH & \textbf{78.23} & 62.5 & \textbf{55.6} & \\
 &\LLMKnn & 67.73  & 60.15 & 55.5 &\\
 
\cline{1-6}
\bottomrule
    \end{tabular}
  \caption[ECB+ Results]{{\conll}~F1 Baseline and Cross-encoder results on \ecb, \ecbM$_1$~and \ecbM$_m$ Test sets. 
	}
\label{tab:ce-ecb-Meta}
\vspace*{-3mm}
\end{table}

\paragraph{CDEC Scores:} We present the overall \conll~F1 scores in Table \ref{tab:ce-ecb-Meta} for the baseline (\LH), the two fine-tuned cross-encoders (\CELH, \CEKnn), and the methods that use GPT-4 (\LLMLH, \LLMKnn). From the table, it is evident that \LH~is no longer a strong baseline for \ecbM~versions with a drop in 20\% score. Both \CELH~and \CEKnn~show a pattern of reducing score from \ecbMone~to \ecbMm, with \CELH~performing considerably worse. Interestingly, the drop in scores for \CEKnn~is not substantial for \ecbMone~but there is a dramatic drop of 20\% for \ecbMm. \LLMLH~achieves the highest scores on \ecb~and \ecbMm, demonstrating that GPT-4's performance aligns with the state-of-the-art, unlike its predecessor GPT-2. However, the financial implications of using \LLMLH~and \LLMKnn~are noteworthy; running CDEC with these methods incurred approximately \$75 in API costs to OpenAI.

From these results, we can conclude three things: a) \ecb~is an easy dataset, b) datasets with complex metaphors are harder benchmarks, and c) GPT-4 is only as good as the \CE~methods with a significant amount of added costs.

\paragraph{Lexical Diversity:} We estimate the lexical diversity (MLTD; \citet{mccarthy2010mtld}) of the mention triggers of event clusters. We first eliminate singleton clusters. Then we calculate a weighted average (by cluster size) of the MLTD score for each cluster. The scores we achieved for the test sets of each version of \ecb~are as follows: ECB+: 7.33. \ecbM1: 11.92, \ecbMm: 26.48. From the lower CDEC scores from Table 2 and the increasing diversity scores of the more complex corpus, we can establish a negative correlation between CDEC scores and MLTD.

Overall, the results confirm our hypothesis that when a dataset a) moves away from strong lexical overlap and b) has figurative language usage, the CDEC scores drop.

\vspace{-0.3mm}
\section{Analysis}
\vspace{-0.3mm}
\subsection{Coreference Resolution Difficulty}
We evaluate whether the paraphrased versions are more difficult for humans to determine as coreferent. On the \devSmall~splits of \ecbMone, \ecbMm, and \ecb, a human annotator reaches the same coreference verdict regardless of the degree of figurative language approximately 98\% of the time. Cases in which the human annotator did not reach the same verdict generally involved convergent metaphorical language, for example: 

\begin{mybox}[]
\small
\textbf{Event a}: The Indian navy unfurled the words that it had ensnared 23 pirates \textbf{\textit{in the law's net}} who cast ominous shadows over a merchant vessel in the Gulf of Aden on Saturday, the latest in a series of recent violent ballets with Somali pirates.

\vspace{1.2mm}
\textbf{Event b}: Indian Naval Ship \textbf{\textit{throws a net over}} three pirate vessels in a single orchestrated symphony .
\end{mybox}

\noindent were incorrectly identified as coreferent; in actuality the former refers to the arrest of the pirates but the latter refers to the interception of their ships.
This analysis supports the findings of \citet{ORTONY1978465}: that, for humans, figurative language use and literal language do not substantially affect comprehension.
\vspace{-0.3mm}
\subsection{Qualitative Error Analysis}
\vspace{-0.1mm}
We examined the coreference predictions of \CEKnn~on 142 common mention pairs between \ecb, \ecbMone, and \ecbMm, as \CEKnn~achieved the best overall performance. For mention pairs that \CEKnn~correctly predicted as coreferent across all versions, we noticed a pattern: the same event trigger was shared in each (see Figure \ref{fig:error1}).

In cases where \CEKnn~got the prediction right on ECB+ but wrong on the META versions, the event triggers in ECB+ were changed to different ones in the META versions (see Figure \ref{fig:error2}). When \CEKnn~incorrectly predicted coreference on ECB+ but correctly predicted it in the META versions, it was because the same triggers in ECB+ were altered to different ones (see Figure \ref{fig:error3}). This further affirms that the model heavily relies on surface triggers for making coreference decisions.
\section{Future Work}
Future research could explore applying more recent CDEC techniques on \ecbM. These techniques could include symbolic grounding, as discussed in \citet{ahmed-etal-2024-x, ahmed-etal-2024-linear-cross}, and event type categorical cross-encoding, as proposed by \citet{otmazgin-etal-2023-lingmess}. Another
outcome of this research is to use CDEC as a text complexity metric \cite{https://doi.org/10.1111/lnc3.12196} of a corpus. We argue that a corpus is more complex if a CDEC algorithm is not able to identify that different explanations of the same event are the same. An interesting line of future work would be to automatically generate an optimally complex CDEC corpus, i.e., a corpus that yields the lowest coreference score.


In this work, we rely on the GPT-4's metaphor list and substitution choice. The only control we have is to make a coherent choice, however, we find ourselves subjected to the unpredictable outputs, colloquially referred to as ``hallucinations'', generated by GPT-4. In the future, we aim to integrate human feedback into the process of metaphor selection and to employ annotated metaphor databases from studies such as \citet{joseph-etal-2023-newsmet}.



 \vspace{-1mm}
\section{Conclusion}
\vspace{-1mm}
In this paper, we introduced \ecbM~a lexically rich variant of \ecb~using constrained metaphoric paraphrasing of the original corpus. We provide hand-corrected event trigger annotations of two versions of \ecbM~differing in the kind of metaphoric transformation using either single words or phrases. We finally provide baseline results using existing SoTA methods on this dataset and show their limitations when there is substantial lexical diversity in the corpus. Through the provided data and methodology, we lay a path forward for future research in Cross-Document Event Coreference Resolution on more challenging datasets.
\section*{Limitations}
The study faced several limitations, including its focus on a single language-English. 
Some experiments were conducted within a small sample space, especially for \devSmall, potentially leading to biased results and limiting the generalizability of the findings. Finally, while the study utilized variations within a single dataset, the reliance on this sole dataset could introduce inherent biases, affecting the broader applicability of the research outcomes.


\textbf{Reproducibility Concern:}
All the coreferencing experiments are reproducible, but the generation of \ecbM~is not. So we may have vastly different results if a new version of \ecbM~is created with the methodology. However, we released all the generated text that came out of our work and the code to run the experiments.

\textbf{LLMs on ECB+. Contamination Concern}
The GPT-4 has likely been contaminated by the test sets of ECB+, i.e., GPT-4 has been pretained on this benchmark. With the recent work involving GPT and ECB+ \cite{yang2022gpt, Ravi2023COMETMRA, ravi-etal-2023-happens}, it seems likely the test set is also been used in the instruction fine-tuning of GPT-4. But we stress the synthesizing of datasets to battle contamination as we do in our work.

\section*{Ethics Statement}
AI-generated text should always be thoroughly scrutinized before being used for any application. In our work, we provide methods to synthesize new versions of the same real articles. This can have unintentional usage in the propagation of disinformation. This work is only intended to be applied to research in broadening the field of event comprehension. Our work carries with it the inherent biases in news articles of ECB+ corpus and has the potential of exaggerating it with the use of GPT-4, which in itself has its own set of risks and biases.
\section*{Acknowledgements}
We thank the anonymous reviewers for their helpful suggestions that improved our paper. We are also grateful to Susan Brown, Alexis Palmer, and Martha Palmer from the BoulderNLP group for their valuable feedback before submission. Thanks also to William Held and Vilém Zouhar for their insightful comments. We gratefully acknowledge the support of  DARPA FA8750-18-2-0016-AIDA – RAMFIS: Representations of vectors and Abstract Meanings for Information Synthesis and a sub-award from RPI on DARPA KAIROS Program No. FA8750-19-2-1004.  Any opinions, findings, conclusions, or recommendations expressed in this material are those of the authors and do not necessarily reflect the views of DARPA or the U.S. government.
\bibliography{anthology,main}
\bibliographystyle{acl_natbib}




\appendix
\section{ECB+ Corpus}
\label{sec:app-ecb}
\renewcommand{\arraystretch}{1.4}
\begin{table}[htb]
\centering
\small
\begin{tabular}{c@{}|ccccc|c@{}}
\toprule
\multicolumn{2}{c}{~} & \textbf{Train} & \textbf{Dev*} & \textbf{\devSmall*} & \multicolumn{1}{c}{\textbf{Test}} & \\
 \hline
& Topics &  25 & 8 & 8& 10 & \\
& Documents &  594 & 156 & 40& 206 & \\
& Mentions &  3808 & 968 & 277& 1780 & \\

\cline{1-7}
\bottomrule
\end{tabular}
\caption[\ecb~Corpus Statistics]{
Corpus statistics for event mentions in ECB+}
\label{tab:ecb}
\end{table}
The ECB+ corpus \cite{cybulska-vossen-2014-using} is a popular English corpus used to train and evaluate systems for event coreference resolution. It extends the Event Coref Bank corpus (ECB; \citet{bejan-harabagiu-2010-unsupervised}), with annotations from around 500 additional documents. The corpus includes annotations of text spans that represent events, as well as information about how those events are related through coreference. We divide the documents from topics 1 to 35 into the training and validation sets\footnote{Validation set includes documents from the topics 2, 5, 12, 18, 21, 34, and 35}, and those from 36 to 45 into the test set, following the approach of \citet{cybulska-vossen-2015-translating}.  We further break the documents of the validation set into two subsets: Dev and \devSmall~for our error analysis. Full corpus statistics can be found in Table \ref{tab:ecb}.


\begin{figure}[t!]
    \centering
    \begin{mybox}[title=\tt Metaphoric Paraphrasing]
    \small
    \tt
    \vspace{1.5mm}
    You are a metaphor expert. Your task is to transform specific words in a given sentence into metaphors. These metaphors can only be \textbf{single-word}/\textbf{multi-word} replacements. Here are the detailed steps you need to follow:

\vspace{1.5mm}
\textbf{Read the Sentence Provided}: Focus on understanding the context and meaning of the sentence.

\textbf{Review the Word List}: This list contains the words you need to transform into metaphors.

\textbf{Generate Metaphors}:

Create \textbf{5 distinct} single-word/multi-word metaphors for each word in the list.

\vspace{2mm}
\textbf{Compose a New Sentence}:

Replace the original words with your chosen metaphors randomly. Ensure the new sentence maintains logical and grammatical coherence.

\textbf{Sentence to Transform}:

"""\{\{sentence\}\}"""

\textbf{Word List to Convert into Metaphors}:

"""\{\{trigger\_list\}\}"""

\vspace{2mm}
\textbf{Output Requirements}:
Provide your final output in JSON format, including:

The "Original Sentence".

The "Original Word List".

The "Metaphoric Word List" (with your chosen metaphors).

The "Metaphoric Sentence" (the sentence with metaphors incorporated).

\vspace{2mm}
Remember, the goal is to use metaphors to convey the original sentence's meaning in a more nuanced or impactful way without altering the core information.
\end{mybox}

    \caption{Metaphoric Paraphrasing Prompt following Chain of Thought Reasoning. We provide the steps in this prompt to follow.}
    \label{fig:meta-prompt}
\end{figure}

\begin{figure}[t]
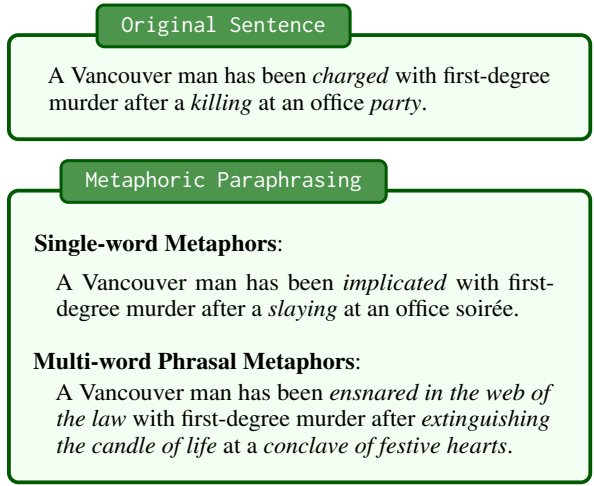

    \centering
    \begin{mybox}[title=\tt Original Sentence]
    \small
    \vspace{1mm}
    A Vancouver man has been \textit{charged} with first-degree murder after a \textit{killing} at an office \textit{party}.
\end{mybox}

\begin{mybox}[title=\tt Metaphoric Paraphrasing]
    \small
    \vspace{1mm}
    \begin{itemize}[itemindent=10.em, leftmargin=1mm, itemsep=1mm]
  \item[\textbf{Single-word Metaphors}:]~

     A Vancouver man has been \textit{implicated} with first-degree murder after a \textit{slaying} at an office soirée.
  \end{itemize}

  \begin{itemize}[itemindent=13.25em, leftmargin=1mm]
      \item[\textbf{Multi-word Phrasal Metaphors}:] ~

      \vspace{-1mm}
      A Vancouver man has been \textit{ensnared in the web of the law} with first-degree murder after \textit{extinguishing the candle of life} at a \textit{conclave of festive hearts}.
  \end{itemize}
  
\end{mybox}

    \caption{Metaphoric Paraphrasing: Transforming a Sentence with Figurative Language. Event triggers, indicated in italics, undergo modification in paraphrased versions, annotated by GPT-4 with two variations. }
    \label{fig:paraphrase}
\end{figure}

\section{Metaphoric Paraphrase Prompt}
\label{sec:app:metaprompt}
We present the prompt used with GPT-4 in Figure \ref{fig:meta-prompt} for generating the Metaphoric Paraphrasing of ECB+ documents. We use two separate prompts for generating single-word metaphors and multi-word metaphors. We ran this prompt on the validation and test sets of ECB+ using GPT-4 as the LLM and a temperature value of 0.7. We force GPT-4 to produce JSON-style output to avoid parsing issues.  It costs about \$16 to generate \ecbMone~and \$18 to generate \ecbMm~with GPT-4 API calls.  In the future, we plan to provide this conversion of the training set of ECB+ as well.

\section{Experiment Setup}
\LH~details: we set the sentence-level word overlap ratio threshold at 0.005. We employ spaCy 3.7.4 as the lemmatizer to extract the root forms of words.

\BEKnn~details: we adopt the RoBERTa-Base model, enhanced with a triplet loss function calculated by \texttt{F.triplet\_margin\_loss} with a 10 margin, L2 norm (\(p=2\)), and \(\epsilon=1e-6\) for stability, without swapping and mean reduction. Our optimization uses AdamW, targeting bi-encoder parameters with a \(1 \times 10^{-5}\) learning rate across 20 iterations and batches of 4.

\CELH~details: We utilize the RoBERTa-Base model with the AdamW optimizer. Learning rates are set to \(1 \times 10^{-5}\) for BERT class parameters and \(1 \times 10^{-4}\) for the classifier. The model is trained over 20 epochs, using the sentences in which the two mentions occur as context, and mention pairs generated by \LH.


\CEKnn~details: It mirrors the \CELH configuration but it is trained on   mention pairs from \BEKnn exclusively.


All Non-GPT experiments are conducted on a single NVIDIA RTX 3090 with 24GB of VRAM. For generating the META datasets, we utilized GPT-4 (model version: gpt4-0613), setting the temperature parameter to 0.7.




\label{sec:app-cross}
\section{\ecbMm~Complete Results}
We provide the baseline results for validation sets of \ecbMm. As shown in Table \ref{tab:ce-ecb-dev}, the results are consistent even for the development sets, where we see significantly low coreference scores with the used methods. Interestingly, \LH~performs better than the cross-encoder methods on these splits.
\begin{table}[t]
\centering
\small
    \begin{tabular}{c@{}|lcccc|c@{}}
        \toprule
        \multicolumn{1}{c}{Split} & \textbf{Method} & \BcuR & \BcuP & \BcuF & \multicolumn{1}{c}{\textbf{\conll}} &  \\ 
        \hline
        & \LH &  51.8&  64.5&  57.4&  56.3& \\
        & \CELH & 47.2 & 77.3 & 58.6 & 55.3 & \\
        \multirow{-3}{*}{Dev}  & \CEKnn & 42.4 & 86.2 & 56.8 & 49.2 & \\
        \cline{1-7}
        & \LH & 68.4 & 78.3 & 73.1 & 62.0 & \\
        & \CELH& 64.8 & 84.7 & 73.4 & 59.0 & \\
        \multirow{-3}{*}{\devSmall}  & \CEKnn & 62.4 & 91.6 & 74.2 & 55.5 & \\
        \cline{1-7}
        \bottomrule
    \end{tabular}
  \caption[ECB+ Results]{Baseline and Cross-encoder results on \ecbMm~Dev and \devSmall sets. 
	}
\label{tab:ce-ecb-dev}
\end{table}

\section{Error Analysis}

\begin{figure}[b!]
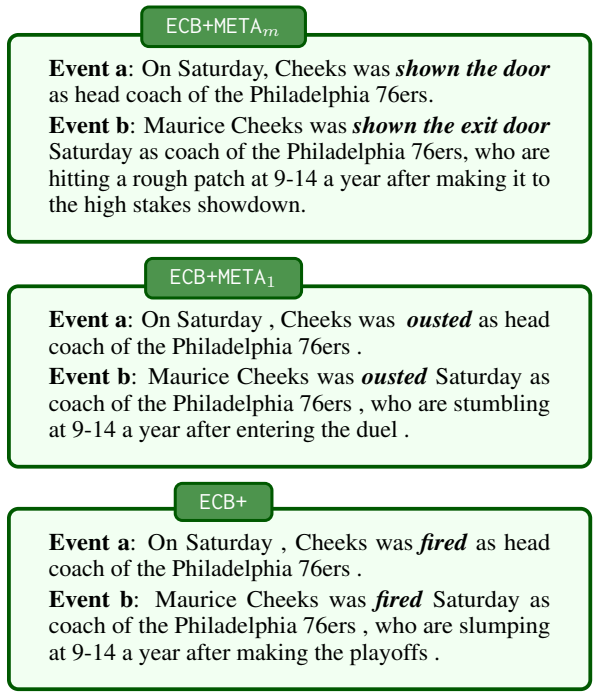

    \centering
    \small
    \begin{mybox}[title=\ecbMm]
        \textbf{Event a}: On Saturday, Cheeks was \textit{\textbf{shown the door}} as head coach of the Philadelphia 76ers.

\vspace{.5mm}
\textbf{Event b}: Maurice Cheeks was \textit{\textbf{shown the exit door}} Saturday as coach of the Philadelphia 76ers, who are hitting a rough patch at 9-14 a year after making it to the high stakes showdown.
    \end{mybox}
    \begin{mybox}[title=\ecbMone]
        \textbf{Event a}: On Saturday , Cheeks was \textit{\textbf{ ousted}} as head coach of the Philadelphia 76ers .

\vspace{.5mm}
\textbf{Event b}: Maurice Cheeks was \textit{\textbf{ousted}} Saturday as coach of the Philadelphia 76ers , who are stumbling at 9-14 a year after entering the duel .
    \end{mybox}
    \begin{mybox}[title=\ecb]
        \textbf{Event a}: On Saturday , Cheeks was  \textit{\textbf{fired}}  as head coach of the Philadelphia 76ers .

\vspace{.5mm}
\textbf{Event b}: Maurice Cheeks was \textit{\textbf{fired}}  Saturday as coach of the Philadelphia 76ers , who are slumping at 9-14 a year after making the playoffs .
    \end{mybox}
    \caption{Correct prediction of coreferent mention pair across all datasets with \CEKnn. Pairs have the same event trigger in each case.}
    \label{fig:error1}
\end{figure}

\begin{figure}[t!]
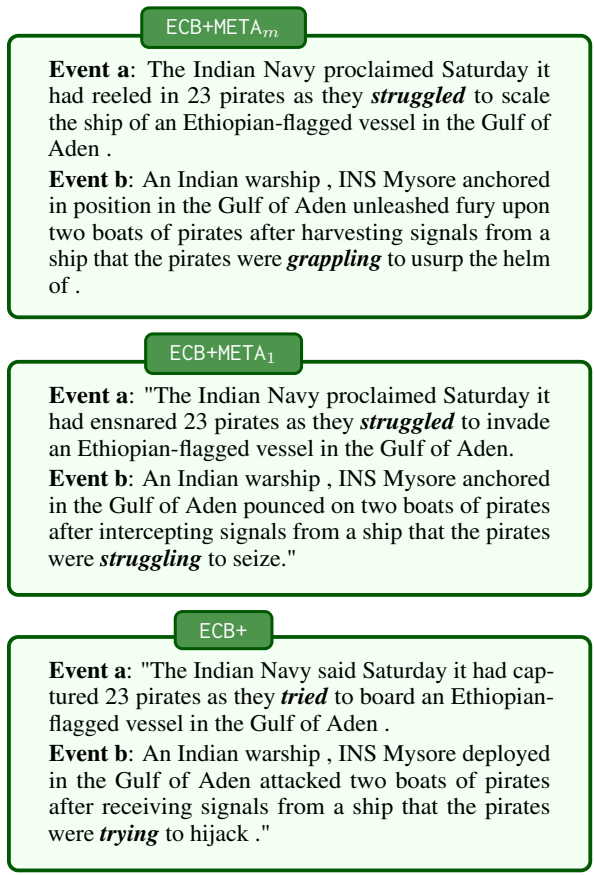

    \centering
    \small
    \begin{mybox}[title=\ecbMm]
        \textbf{Event a}: The Indian Navy proclaimed Saturday it had reeled in 23 pirates as they \textit{\textbf{struggled}} to scale the ship of an Ethiopian-flagged vessel in the Gulf of Aden .

\vspace{0.5mm}
\textbf{Event b}: An Indian warship , INS Mysore anchored in position in the Gulf of Aden unleashed fury upon two boats of pirates after harvesting signals from a ship that the pirates were  \textit{\textbf{grappling}}  to usurp the helm of .
    \end{mybox}
    \begin{mybox}[title=\ecbMone]
        \textbf{Event a}: "The Indian Navy proclaimed Saturday it had ensnared 23 pirates as they  \textit{\textbf{struggled}}  to invade an Ethiopian-flagged vessel in the Gulf of Aden.

\vspace{0.5mm}
\textbf{Event b}: An Indian warship , INS Mysore anchored in the Gulf of Aden pounced on two boats of pirates after intercepting signals from a ship that the pirates were  \textbf{\textit{struggling}}  to seize."
    \end{mybox}
    \begin{mybox}[title=\ecb]
        \textbf{Event a}: "The Indian Navy said Saturday it had captured 23 pirates as they  \textit{\textbf{tried}}  to board an Ethiopian-flagged vessel in the Gulf of Aden .

\vspace{0.5mm}
\textbf{Event b}: An Indian warship , INS Mysore deployed in the Gulf of Aden attacked two boats of pirates after receiving signals from a ship that the pirates were  \textit{\textbf{trying}}  to hijack ."
    \end{mybox}
    \caption{Correct coreference prediction in \ecb~but not in the META versions, simply because the triggers got changed.}
    \label{fig:error2}
\end{figure}

\begin{figure}[t!]
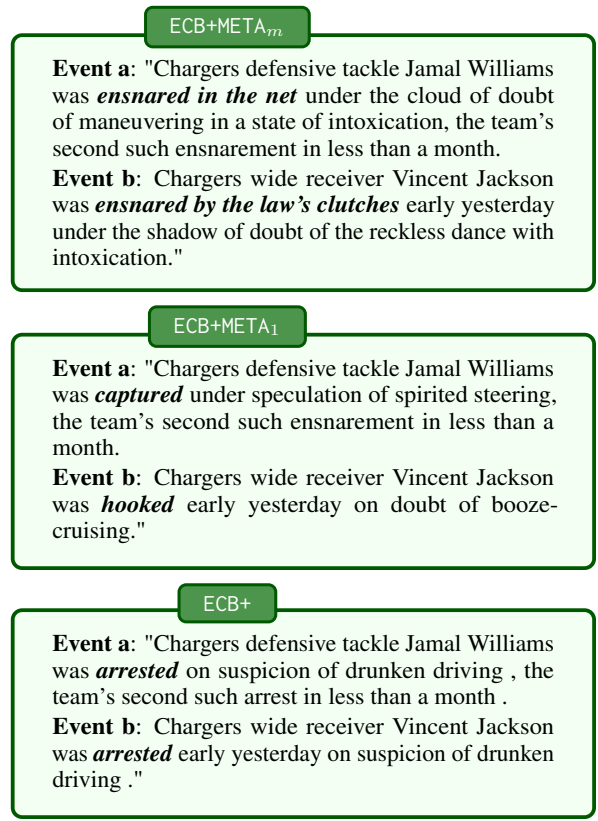

    \centering
    \small
    \begin{mybox}[title=\ecbMm]
        \textbf{Event a}: "Chargers defensive tackle Jamal Williams was  \textit{\textbf{ensnared in the net}}  under the cloud of doubt of maneuvering in a state of intoxication, the team's second such ensnarement in less than a month.

\vspace{0.5mm}
\textbf{Event b}: Chargers wide receiver Vincent Jackson was  \textit{\textbf{ensnared by the law's clutches}}  early yesterday under the shadow of doubt of the reckless dance with intoxication."
    \end{mybox}
    \begin{mybox}[title=\ecbMone]
        \textbf{Event a}: "Chargers defensive tackle Jamal Williams was  \textit{\textbf{captured}}  under speculation of spirited steering, the team's second such ensnarement in less than a month.

\vspace{0.5mm}
\textbf{Event b}: Chargers wide receiver Vincent Jackson was  \textit{\textbf{hooked}}  early yesterday on doubt of booze-cruising."
    \end{mybox}
    \begin{mybox}[title=\ecb]
\textbf{Event a}: "Chargers defensive tackle Jamal Williams was  \textit{\textbf{arrested}}  on suspicion of drunken driving , the team's second such arrest in less than a month .

\vspace{0.5mm}
\textbf{Event b}: Chargers wide receiver Vincent Jackson was  \textit{\textbf{arrested}}  early yesterday on suspicion of drunken driving ."
    \end{mybox}
    \caption{Correct non-coreference prediction in \ecbM~but not in \ecb, simply because the META versions' event triggers were changed.}
    \label{fig:error3}
\end{figure}

For more examples, please checkout the provided excel file in data repository.

\label{sec:app-error}
\end{document}